\pdfoutput=1

\documentclass[11pt]{article}

\usepackage{naacl2021}

\usepackage{times}
\usepackage{latexsym}

\usepackage[T1]{fontenc}

\usepackage[utf8]{inputenc}

\usepackage{microtype}

\usepackage{url}
\usepackage{subfigure}
\usepackage{amsfonts} 
\usepackage{graphicx}
\usepackage{tabularx}
\usepackage{bigstrut}
\usepackage{multirow}
\usepackage{color}

\title{GroupLink: An End-to-end Multitask Method for Word Grouping and Relation Extraction in Form Understanding}

\author{
Zilong Wang\textsuperscript{\dag} \quad 
Mingjie Zhan\textsuperscript{*} \quad 
Houxing Ren\textsuperscript{\ddag} \quad 
Zhaohui Hou\textsuperscript{*} \\ 
{\bf Yuwei Wu}\textsuperscript{*} \quad 
{\bf Xingyan Zhang}\textsuperscript{*} \quad 
{\bf Ding Liang}\textsuperscript{*} \\ 

\textsuperscript{\dag}University of California, San Diego \\
\textsuperscript{*}Sensetime Group Limited \\
\textsuperscript{\ddag}Beihang University \\
\texttt{zlwang@ucsd.edu, \{zhanmingjie, houzhaohui, wuyuwei, zhangxingyan, }\\\texttt{ liangding\}@sensetime.com, renhouxing@buaa.edu.cn}

}

\begin{document}
\maketitle
\begin{abstract}
  Forms are a common type of document in real life and carry rich information through textual contents and the organizational structure. To realize automatic processing of forms, word grouping and relation extraction are two fundamental and crucial steps after preliminary processing of optical character reader (OCR). Word grouping is to aggregate words that belong to the same semantic entity, and relation extraction is to predict the links between semantic entities. Existing works treat them as two individual tasks, but these two tasks are correlated and can reinforce each other. The grouping process will refine the integrated representation of the corresponding entity, and the linking process will give feedback to the grouping performance. For this purpose, we acquire multimodal features from both textual data and layout information and build an end-to-end model through multitask training to combine word grouping and relation extraction to enhance performance on each task. We validate our proposed method on a real-world, fully-annotated, noisy-scanned benchmark, FUNSD, and extensive experiments demonstrate the effectiveness of our method.
\end{abstract}

\section{Introduction}
Forms are a ubiquitous document format in real life because it carries rich information efficiently through textual contents and layout structure. Countless forms are used in modern finance, insurance, and medical industry every day. Although these forms are electronically stored in formats, e,g, PDF, PNG, it is still necessary to ask massive human labor to organize them and collect valuable information. It is not easy for computers to automatically understand the scanned forms since it remains unstructured data. Modern optical character reader (OCR) is used to have a glimpse at textual contents. However, even if all textual contents are correctly recognized, there is still non-trivial to convert the forms into structured information. The biggest challenge is that the OCR only provides the recognition results of individual words or characters and cannot extract the structural information.

There are two major challenges to bridge the gap between the OCR results and the structural information: the word grouping and relation extraction. These two tasks solve the issues on textual contents understanding and layout structure analysis, respectively. To be more specific, the two tasks are clearly defined as follows.

\begin{figure}
    \centering
    \includegraphics[width=0.9\linewidth]{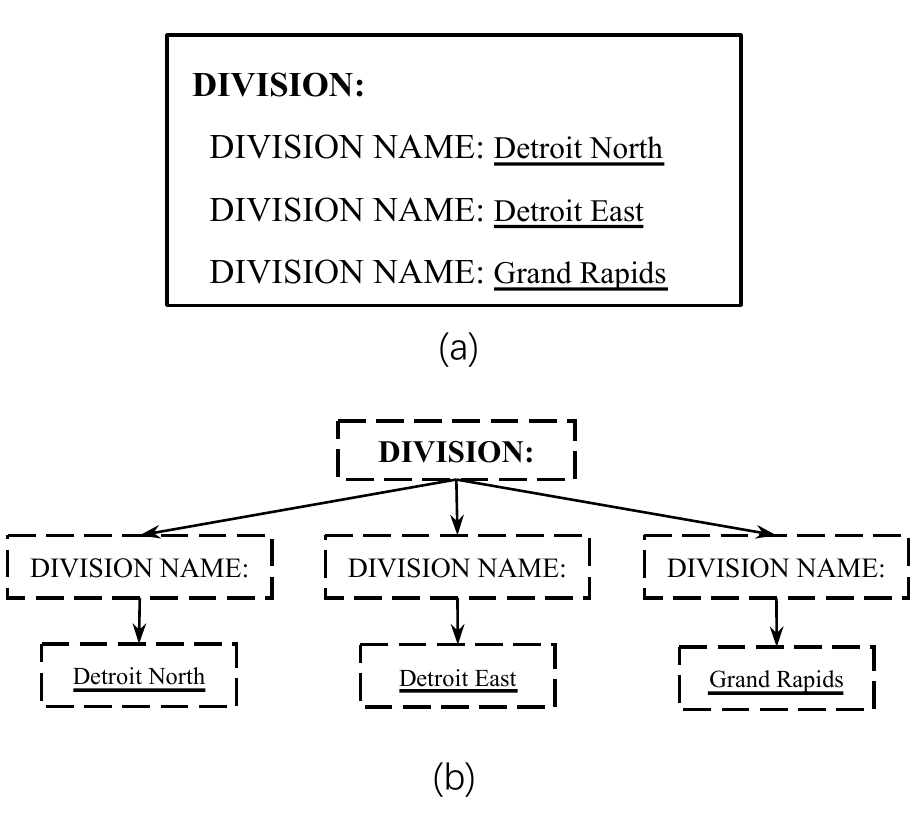}
    \caption{Relation Extraction Example: (a) is part from a form in dataset FUNSD. (b) is the annotated links among them. The arrays between the entities denote the relation links.}
    \label{fig:link_example}
\end{figure}

\paragraph{Word grouping:} aggregate words that belong to the same semantic entity. 
As mentioned above, modern OCR techniques have already achieved satisfactory performance in recognizing textual contents. However, the results are mostly discrete bounding boxes containing single words. Single words are ambiguous to express a concept. Therefore, it is necessary to group words into an integrated semantic entity to convey complete meanings.

\paragraph{Relation Extraction:} predict the links between semantic entities.
An example from our selected dataset is shown in Figure \ref{fig:link_example} \footnote{The figures in this paper are processed according to the original images for clearness.}.
From the example above, we can observe that each semantic entity is integrated and correlated with other entities in the same form page. The most common relation type is key-value pairs. There are three pairs of ``Division name'' and its exact value in the example.

To summarize, the word grouping task guarantees the completeness of each semantic entity and enables computers to understand the exact textual contents in forms; the relation extraction task is built on the basis of the previous task and tries to figure out the layout patterns of semantic entities to extract structural information, such as key-value pairs in forms.

Since these two tasks solve significant issues of form understandings, many works have been proposed. However, they treat them as two individual tasks. We notice that when grouping words into semantic entities, the integrated meaning of the given entity will be more clearly conveyed, and the process will refine the representation of the entity's semantic meaning. On the other hand, the performance of relation prediction will also give feedback to the word grouping. The two tasks are not completely separated but can be reinforced \cite{zhang2020trie}.

Bearing this in mind, we propose an end-to-end method combining the previous two form understanding tasks, word grouping and relation extraction, together. Since forms contain both textual data and layout information, we adopt BERT language model \cite{devlin2018bert} and a layout analysis model to obtain an informative representation of each word in the form page. In the word grouping part, it is easy to notice the partial order within a form page, i.e., the words belonging to the same semantic entity locate nearby and can be sorted according to geometric coordinates. Following \citep{xu2020layoutlm}'s solution on the named entity recognition (NER), we sort the bounding boxes containing single words in a sequence and adopt sequential tagging techniques. Specifically, we use bidirectional long short term memory networks (BiLSTM) followed by conditional random fields (CRF) to segment the semantic entity sequence. Each span serves as an integrated semantic entity. After the word grouping information is acquired in the relation extraction part, we input the word representation features in each span to a multi-layer transformer encoder \cite{vaswani2017attention}. To model the relation between semantic entities, such as key-value pairs, we design an asymmetric relation extraction module and negative sampling to train our model. We validate our end-to-end method on a real-world, fully-annotated, noisy-scanned benchmark, FUNSD \cite{jaume2019}. The benchmark contains various informative forms in different formats. Extensive experiments demonstrate the effectiveness of our proposed method.

We summarize our contribution as follows:
\begin{itemize}
    \item We propose an end-to-end method combining two related form understanding tasks, word grouping and relation extraction, together. As far as we know, we are the first to explore the collaboration between these two tasks.
    \item We extract features from both textual data and layout information to make full use of features available in a form page.
    \item We have achieved state-of-the-art performance on both tasks compared to our baselines.
\end{itemize}

\section{Related Work}

\subsection{Word Grouping}
Word grouping used to be a common computer vision task. Mostly it solves scattered words or characters in some daily scenes, such as billboards of shops or tags in the photos \cite{shahab2011icdar,karatzas2013icdar}. These scenarios are complicated, and it is extremely challenging to find a reading order by sorting. Therefore, most methods use clustering or network techniques. \citep{zhang2020deep} uses relational reasoning graph network to solve the issues of the arbitrarily scattered text. They model the connection between bounding boxes as edges in a network. Through graph neural networks or other techniques, they would like to cluster the words together and produce a meaningful sequence. On the other hand, it is intuitive to adopt natural language knowledge to improve grouping performance since the final goal is to get an entity or phrase that obeys human language laws. \citep{wenhai2020ae} does the first attempt to insert natural language processing into this task. However, the NLP knowledge in their work is used to refine the result. They use computer vision models to extract the texts and then use a pretrained language model to judge whether it is correct.

Unlike the daily scenes, words or characters in forms distribute in some partial order within a page and do not suffer from filming angle or image rotation issues. Although words are sparse in forms and most areas are blank, there are ordered word sequences if we focus on some parts of a form. That is why we can adopt sequence labeling techniques on these forms. In \citep{xu2020layoutlm}, researchers use sequence labeling to solve named entity recognition (NER) in forms. Locating intervals is a sub-task of NER. In forms with partial order, intervals between entities are the only thing we need to group words. Therefore, we propose a sequence labeling model based on long short term memory networks and conditional random fields \cite{huang2015bidirectional,dong2016character} to solve word grouping in forms.

\subsection{Relation Extraction}
Relation extraction in forms is to find the links between semantic entities in a form to explore the structural information. The most common linking relation is key-value pairs. This task is first defined in \citep{jaume2019funsd} as a task of form understanding. Forms are not regular documents, both textual and layout information matters. It is intuitive to require more computer techniques to extract the relation links to save human labor.

In \citep{jaume2019funsd}, they proposed a model use classify whether there is a link between any pair of semantic entities. However, if the problem is set in this way, there will be a huge gap between the number of positive and negative cases. The negative case number will much more than the positive. The performance is not satisfactory, as a result. To eliminate the problem, we use negative sampling in our training process. The details are included in the following sections.

\begin{figure*}[t]
  \centering
  \includegraphics[width=0.75\textwidth]{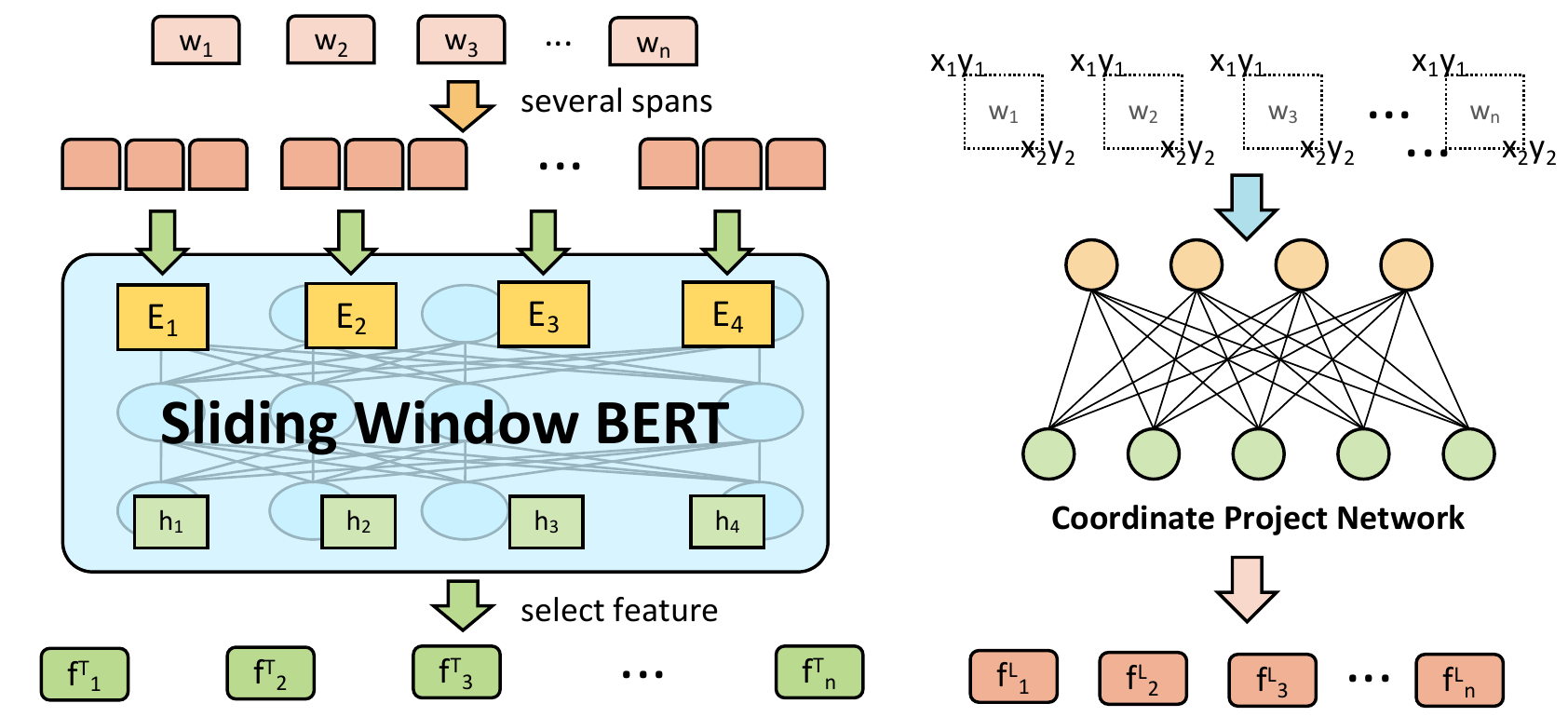}
  \caption{The Text-layout Base Model}
  \label{fig:backbone}
\end{figure*}


\section{Methodology}

In this section, we will first describe the preliminary processing step and introduce an overview of our method. Then we introduce the components of our proposed end-to-end model, \textit{GroupLink}, in details, including the text-layout base model, sequence labeling model for word grouping, and link prediction model for relation extraction. 


\subsection{Overview}
Given a form page, modern optical character recognition (OCR) can provide us with a sequence of bounding boxes containing a single word along with the geometric coordinates. We aim at making full use of these textual data and layout information to group the words and extract the links between them.

Intuitively, textual data are not the only information carrier in a form page. The most significant difference between ordinary documents and form pages is that the layout patterns in forms also provide rich information. That is why we not only use the pretrained language model, BERT \cite{devlin2018bert}, in our feature extraction model, but also includes the layout analysis model to extract layout features from geometric coordinates. Once the text-layout base model extracts the features, the features will be passed to the two substantial models to solve the following form understanding tasks. In the sequence labeling model for word grouping, we follow existing works and apply a bidirectional long short term memory network (BiLSTM) along with a conditional random field (CRF) to segment the sorted sequence and produce semantic entities in each span. Consequently, the link prediction model for relation extraction is equipped with the text-layout features of each word and semantic entity spans of several words. We further input each span of word features into a multi-layer transformer to extract the informative representation of each semantic entity. Finally, we adopt a negative sampling training method to predict the links between these entities.

We denote a form page $F$ as a sequence of bounding boxes: $\{B_i\}_n$, where $B_i$ is a bounding box containing a single word and $n$ is the number of bounding boxes on this page. For bounding box $B_i$, preliminary processing provides us with the textual content, word $w_i$, and the layout information, geometric coordinates $(x^i_1, y^i_1, x^i_2, y^i_2)$. The segmentation label sequence is denoted as $[y_1,...,y_n]$. Supposing that bounding boxes $i$ to $j$ compose of a semantic entity, we denote the entity as $E_{ij}$ or $E^I$, where $I$ is the index of the entity in the form page. The link between entity $I$ and $J$ is denoted as $E^I \to E^J$.

\subsection{Text-layout Base Model}
We aim at extracting both textual features and layout features from a form page through Sliding Window BERT and Coordinate Projection Network, respectively. The final representation of the given bounding box is the concatenation of these two features.

\paragraph{Sliding Window BERT}
Textual features contain much contextual information and semantic signals. Following many existing works, we adopt BERT \cite{devlin2018bert} to acquire them. However, some related works \cite{beltagy2020longformer,zaheer2020big} have pointed out that the BERT language model suffers from the limitation of sequence length. Since the most important part in BERT is the self-attention layer of transformer \cite{vaswani2017attention}, and it is not sensitive to position, the BERT language model has to add trainable positional embeddings of fixed length, which bring difficulties to extend the usage on longer sequences in some scenarios. We use a sliding window strategy to extend the pretrained language model of fixed length to longer sequences and keep the information loss as little as possible.

First, we use a sliding window to separate the sequence of bounding box in a form page into several spans. The length of each span $l$ equals to the max positional embedding length of the BERT. The sliding stride $\delta$ is a hyper-parameter. 
\begin{equation}
  F = [B_1, B_2, ..., B_n]
\end{equation}
\begin{equation}
  span_i = [B_{i\delta},...,B_{i\delta + l - 1}]
\end{equation}

Through this sliding window strategy, each bounding box will exist in multiple spans, and the word in each bounding box will correspond to multiple features extracted from BERT based on different contexts in each span. To select the most informative feature, we need to build a map function $g$ from the word $w_i$ to a selected span $span_j$, and use the BERT result of this word in this specific span as its textual feature.

We apply a min-max algorithm based on the word position in each span to realize it. Given a word $w_i$ and a span $span_j$ containing it, we calculate the distance of the word to the span head and tail. We would like to minimize the maximum of this two distances so as to ensure that the word in the selected span can have the richest context.
\begin{equation}
  g(w_i) = \mbox{}{argmin}_{j, B_i\in span_j} \mbox{max}(head, tail)
\end{equation}
where $g(w_i)$ is the index of the selected span; $head$, $tail$ are the word's distance to the span head and tail, which can be calculated through $i - j\delta$, $j\delta + l - 1 - i$, respectively..

Then, we input words in the selected span $span_{g(w_i)}$ into a BERT language model to produce the required textual feature $f^{TX}_i$.
\begin{equation}
  f^{TX}_i = \mbox{BERT}(span_{g(w_i)})[w_i]
\end{equation}

\begin{figure*}[t]
  \centering
  \includegraphics[width=0.9\textwidth]{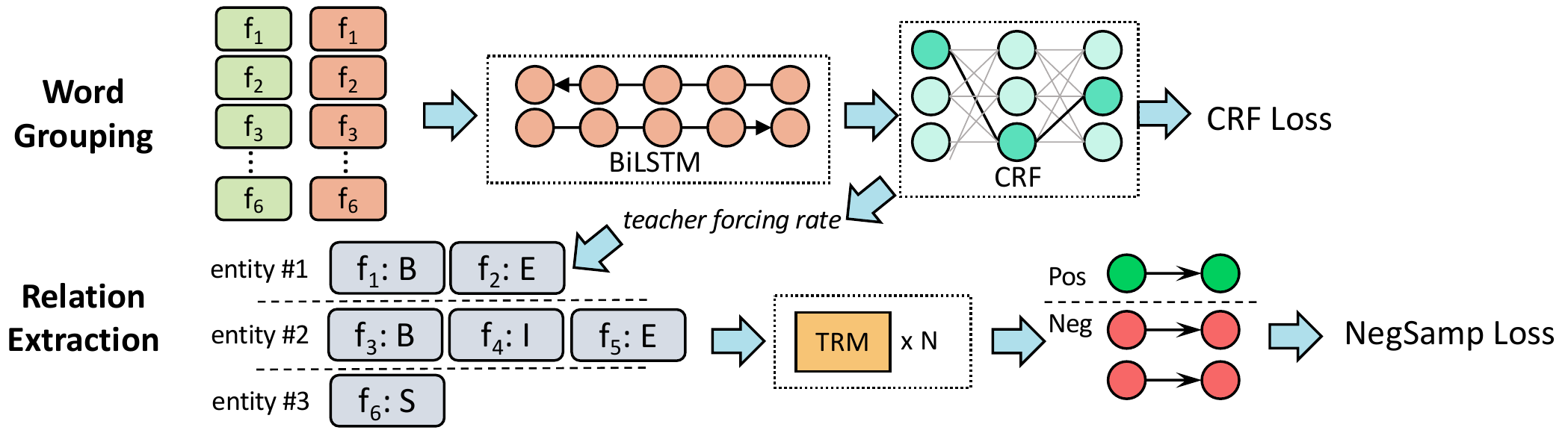}
  \caption{The Sequence Labeling Model for Word Grouping task and the Link Prediction Model for Relation Extraction Task}
  \label{fig:multitask}
\end{figure*}

\paragraph{Coordinate Projection Network}
Different from textual data, the layout information, i.e. geometric coordinates, is difficult to use. Textual data convey rich information and modern natural language processing techniques have expressed the essential meaning in a fixed dimension of vector. In comparison, coordinates are poor in information capacity. In this paper, we only have the 4-dimension coordinates (left-top and right-bottom) of a bounding box. To extract the latent features from the coordinates, we adopt a simple and useful model to project the coordinates to a hyperspace so as to obtain the layout features $f^L_i$.
\begin{equation}
  geo_i = [x^i_1, y^i_1, x^i_2, y^i_2]    
\end{equation}
\begin{equation}
  f^{LT}_i = \sigma(W\cdot geo_i + b)
\end{equation}
where $\sigma$ is activate function and we use $relu(\cdot)$ here; the $W$ and $b$ are trainable parameters of multi-layer perceptron.

Finally, we take both features into account and consider both textual data and layout information. We produce the representation of a bounding box through the concatenation of the textual feature and layout feature.
\begin{equation}
  f_i = [f^{TX}_i;f^{LT}_i]
\end{equation}
where the $(;)$ is the operation of concatenation.

\subsection{Sequence Labeling Model}
Considering bounding boxes have partial order, we apply segmentation algorithms on the sequence of bounding boxes to find the interval of each semantic entities. To be specific, following \citep{xu2020layoutlm}'s solution on named entity recognition, we use tags \verb|B|, \verb|I|, \verb|E| to denote the beginning, inside, ending word of a multi-word entity, and we use tag \verb|S| to denote single-word entity.

We use a sequence labeling model to predict the tag of each word. Our model is composed of bidirectional long short term memory network (BiLSTM) and conditional random field (CRF) \cite{huang2015bidirectional,dong2016character}. The BiLSTM will further extract the contextual features, and it is beneficial to consider the correlation between labels of neighboring words. The prediction of labels will be decoded as the most probable chain of labels using the Viterbi algorithm.

After the text-layout base model, we have extracted a sequence of features. Given a possible label sequence $\tilde{y} = [\tilde{y}_1,\tilde{y}_2,...,\tilde{y}_n]$, we input the features to BiLSTM. The CRF takes the hidden states of BiLSTM $h$ as inputs and model the probability of labels $\tilde{y}$ given the observing features as follows.
\begin{equation}
  h = \mbox{BiLSTM}([f_1,f_2,...,f_n])
\end{equation}
\begin{equation}
  p(\tilde{y}|h,\theta) = \frac{\prod_i \zeta_i(\tilde{y}_i,\tilde{y}_{i+1},h)}{Z(h)}
\end{equation}
where $\theta$ is a set of trainable parameters; $Z(h)$ is normalization factor; $\zeta_i(y,y^\prime,h)$ is potential function to model the correlation of label $y$ and $y^\prime$ at step $i$.

We use maximum conditional likelihood estimation to train the CRF. The loss function is calculated as follows.
\begin{equation}
  \mathcal{L}_{CRF} = -\sum \mbox{log}p(y|h, \theta)
\end{equation}
where $y$ is the correct label sequence.

The Viterbi algorithm will decode the most probable chain of labels as prediction result $\hat{y}$. The accuracy between the correct labels $y$ and predicted labels $\hat{y}$ is used as metric to evaluate the performance.
\begin{equation}
  \hat{y} = \mbox{argmax}_{\tilde{y}}\mbox{log}p(\tilde{y}|h, \theta)
\end{equation}

\subsection{Link Prediction Model}
The link prediction model is designed to predict the correlation link between two semantic entities to extract the relation between them and use a scalar to evaluate the probability. We separate this task into three steps, entity feature extraction, entity relation modeling, and negative training.

\paragraph{Entity Feature Extraction}
Although each semantic entity can be collected through the segmentation labels from the word grouping task, the text-layout features from the base model are still word-wise. We use the same multi-layer transformer to encode all entities' text-layout feature sequences to obtain the representation of a semantic entity. Besides, following the idea of classification tag in BERT, we add two unique trainable embedding vectors $\alpha, \beta \in \mathbb{R}^d$, where $d$ is the dimension of text-layout features, at the head and tail of each span, corresponding to \verb|[CLS]| and \verb|[SEP]| in BERT input sequence.

Supposing that entity $E^I$ is composed of bounding boxes from index $i$ to $j$, then its representation is calculated as follows.
\begin{equation}
  E^I = E_{ij} = [B_i,B_{i+1}, ..., B_j]
\end{equation}
\begin{equation}
  F^I = \mbox{TRM}^{m}([\alpha, f_i, f_{i+1}, ..., f_{j}, \beta])[\alpha]
\end{equation}
where $f_i$ is the text-layout feature corresponding to the bounding box $B_i$; $\mbox{TRM}^m$ is the multi-layer transformer and $m$ is the number of layers; $F^I$ is the semantic entity $E^I$'s feature and it is extracted from the last layer's hidden states of trainable vector $\alpha$.

\paragraph{Entity Relation Modeling}
After the extraction of entity features, we need to model the entity relation. The links between semantic entities we care about are asymmetric. The direction of relation plays an important role in this scenario. Therefore, some common symmetric relation metrics, e.g., Dot Production, Euclidean distance, Poincar\'e distance \cite{nickel2017poincare}, are no longer feasible here.

To model the links between semantic entities, the function $\mathcal{P}_{E^I\to E^J}$ needs to meet the requirement that: $\mathcal{P}_{E^I\to E^J} \not= \mathcal{P}_{E^J\to E^I}$. We utilize a parameter matrix to model the linking relation. We put the matrix in the middle of two entity features. Given the two entities $E^I$ and $E^J$, the probability of $E^I \to E^J$ is calculated through:
\begin{equation}
  \mathcal{P}_{E^I \to E^J} = F^I M (F^J)^T \in \mathbb{R} \label{eq:relation}
\end{equation}
where $M$ is an asymmetric parameter matrix, so $\mathcal{P}_{E^I\to E^J} \not= \mathcal{P}_{E^J\to E^I}$; $F^I$ and $F^J$ are entity features of entity $E^I$ and $E^J$, respectively.

\paragraph{Negative Sampling}
It should be noted that links do not always exist between entities. A randomly selected pair of semantic entities are more likely to be unrelated at all. Therefore, the training meets the problem of data sparsity. To balance the different number of linked entity pairs and unlinked ones, we adopt the negative sampling \cite{mikolov2013distributed} to train our model.

According to human annotation, we have got several positive cases of linked entities. For each case $E^I \to E^J$, we randomly select a fixed number of unrelated entities for $E^J$. Our model is supposed to distinguish the correct entity from the false ones. More specifically, we denote the set of negative examples as $N$. The $E^K\in N$ are not linked to $E^J$.


We calculate the negative sample loss as follows. We normalize the probability and minimize the cross entropy loss of $\mathcal{P}_{E^I\to E^J}$. So we can maximize the $\mathcal{P}_{E^I\to E^J}$ and minimize the $\mathcal{P}_{E^I\to E^K}$.
\begin{equation}
  \mathcal{L}_{Neg} = -\log \sum_{E^I \to E^J} \frac{e^{\mathcal{P}_{E^I\to E^J}}}{e^{\mathcal{P}_{E^I\to E^J}}+\sum_{E^K \in N} e^{\mathcal{P}_{E^K\to E^J}}}
\end{equation}
where $N$ is the negative sampling set of semantic entity $E^J$ and $\mathcal{P}_{E^I\to E^J}$ is the probability of a link existing between semantic entity $E^I$ and $E^J$.

\subsection{Multitask Training}
The two form understanding tasks mentioned in previous sections are correlated and can reinforce each other. We aim at improving the performance of each task through multitask training. Inspired by teacher forcing rate in image detection and image recognition tasks, we adopt the teacher forcing rate while multitask training. In the link prediction model, the segmentation results are necessary to collect semantic entities for relation extraction. The segmentation result has two sources. The link prediction model can take the results from the sequence labeling model or the correct labels from the human annotation. The teacher forcing rate in our model is to decide the segmentation label source. In this way, the link prediction model will give feedback on the performance of word grouping.

In the final optimization step, the multitask training loss is the sum of CRF loss in sequence label model and negative sampling loss in link prediction model.
\begin{equation}
  \mathcal{L} = \mathcal{L}_{CRF} + \mathcal{L}_{Neg}
\end{equation} 

\section{Experiment}

In this section, we conduct experiments on benchmark FUNSD \cite{jaume2019funsd} \footnote{The FUNSD dataset is available at \url{https://guillaumejaume.github.io/FUNSD/}} to validate the effectiveness of our proposed end-to-end method in three scenarios: only word grouping task training, only relation extraction task training, and both tasks joint training. It is noted that only the loss function corresponding to the task is included when optimization in the single-task training scenario.

\subsection{Dataset}
We select dataset FUNSD \cite{jaume2019funsd} as our benchmark. FUNSD is composed of real, fully annotated, scanned forms. The forms are noisy and vary widely in appearance, making form understanding a challenging task. The annotation has two parts: the first part provides the semantic entities in several neighboring bounding boxes and the geometric coordinates; the second part provides the relation links. The statistics are listed in Table \ref{tab:funsd}.

\begin{table}[htbp]
  \centering
  \caption{Statistics of dataset FUNSD}
    \begin{tabular}{|c|c|c|c|c|}
    \hline
          & \textbf{Forms} & \textbf{Boxes} & \textbf{Entities} & \textbf{Links} \bigstrut\\
    \hline
    train & 149   & 21888 & 7259  & 4154 \bigstrut\\
    \hline
    test  & 50    & 8707  & 2270  & 1057 \bigstrut\\
    \hline
    total & 199   & 30595 & 9529  & 5211 \bigstrut\\
    \hline
    \end{tabular}%
  \label{tab:funsd}%
\end{table}%


\subsection{Word Grouping Task Comparison}
Following \citep{xu2020layoutlm}'s solution on named entity recognition, we utilize the partial order of the bounding boxes and sort the bounding boxes into a sequence. Our job is to find the intervals of entities in the sequence. In this section, we replace the base model and classifier with several alternatives as comparative baselines to prove the effectiveness of our proposed model on the word grouping task. The accuracy is used as metrics.

The alternative of feature extraction base model:
\textbf{BERT} \cite{devlin2018bert}: BERT is outstanding in extracting textual features, but the layout information is not considered in ordinary BERT.
\textbf{LayoutLM} \cite{xu2020layoutlm}: LayoutLM is the first pretrained model using both textual data and layout information. It outperforms ordinary BERT on many document understanding tasks. It leverages the coordinates in the embedding layer to merge the layout features into the language model.

The alternative of classifier: \textbf{Multi-layer Perceptron (MLP)}: MLP is a well-known model for classification tasks. It utilizes linear transformation and non-linear activation function. 

We combine the base models and the classifiers and conduct experiments in the same environment settings. The results are in Table \ref{tab:word_group_result} 

\begin{table}[htbp]
  \centering
  \caption{Comparison Result of Word Grouping Task}
    \begin{tabular}{|l|l|c|}
    \hline
    \textbf{Model} & \textbf{Classifier} & \textbf{Accuracy} \bigstrut\\
    \hline
    \multirow{2}[4]{*}{BERT} & MLP & 79.75 \bigstrut\\
\cline{2-3}          & BiLSTM + CRF & 88.07 \bigstrut\\
    \hline
    \multirow{2}[4]{*}{LayoutLM} & MLP & 79.25 \bigstrut\\
\cline{2-3}          & BiLSTM + CRF & 90.56 \bigstrut\\
    \hline
    \multirow{2}[4]{*}{\textbf{TL base model}} & MLP & 80.78 \bigstrut\\
\cline{2-3}          & \textbf{BiLSTM + CRF} & \textbf{91.55} \bigstrut\\
    \hline
    \end{tabular}%
  \label{tab:word_group_result}%
\end{table}%

From the Table \ref{tab:word_group_result}, we observe that our proposed sequence labeling model for word grouping task achieves best performance over other alternatives. In the comparison of base models, we can see the layout information contributes to the labeling results. A clear improvement can be seen when we add layout features. It is reasonable because the intervals of entities are closely related to the geometric position. Compared with LayoutLM, our proposed text-layout base model uses layout information differently and guarantees the continuity of coordinate features, eliminating gaps between neighboring coordinates. Moreover, in the comparison of classifiers, the combination of BiLSTM and CRF performs much better than Multi-layer Perceptron. It is not surprising because the segmentation labels are highly correlated, and CRF is good at modeling the transform patterns between labels. The carefully designed base model and classifiers provide us with satisfactory results on the word grouping task.

\begin{table*}[htbp]
  \centering
  \caption{Comparison Result of Relation Extraction Task}
    \begin{tabular}{|l|l|c|c|c|c|c|}
    \hline
    \multirow{2}[4]{*}{\textbf{Inputs}} & \multirow{2}[4]{*}{\textbf{Feature}} & \multicolumn{2}{c|}{\textbf{Reconstruction}} & \multicolumn{3}{c|}{\textbf{Detection}} \bigstrut\\
\cline{3-7}          &       & \textbf{mAP} & \textbf{mRank} & \textbf{Hit@1} & \textbf{Hit@2} & \textbf{Hit@5} \bigstrut\\
    \hline
    entities & FUNSD-base & 0.2385 & 11.68 & 10.12 & 16.26 & 36.20 \bigstrut[t]\\
    entities & LayoutLM & 0.4761 & 7.11  & 32.43 & 45.56 & 66.41 \\
    entities & DocStruct & 0.7177 & 2.89  & 58.19 & 76.27 & 88.94 \bigstrut[b]\\
    \hline
    words & \textbf{Ours} & \textbf{0.8297} & \textbf{1.85} & \textbf{72.82} & \textbf{85.33} & \textbf{95.55} \bigstrut\\
    \hline
    \end{tabular}%
  \label{tab:entity_link_result}%
\end{table*}%

\begin{table*}[h]
  \centering
  \caption{Comparison Result of Single-task and Multitask Training of Relation Extraction Task}
    \begin{tabular}{|l|c|c|c|c|c|}
    \hline
    \textbf{Relation Extraction Task} & \textbf{mAP} & \textbf{mRank} & \textbf{Hit@1} & \textbf{Hit@2} & \textbf{Hit@5} \bigstrut\\
    \hline
    single-task training & 0.8297 & 1.85  & 72.82 & 85.33 & \textbf{95.55} \bigstrut\\
    \hline
    multitask training & \textbf{0.8359} & \textbf{1.77} & \textbf{74.15} & \textbf{86.87} & 95.37 \bigstrut\\
    \hline
    \end{tabular}%
  \label{tab:multitask_result_2}%
\end{table*}%

\subsection{Relation Extraction Task Comparison}
Form understanding is to recognize the semantic entities through word grouping task and to extract rich information from the layout patterns through relation extraction task. In our problem setting, we predict the probability of each related entity pair. We design two tasks to evaluate our proposed model's performance.
\paragraph{Reconstruction:} Given the human-labeled linked entity pairs in a form, we predict the key for entities to reconstruct the relation network. The Mean Average Precision (mAP) and Mean Rank (mRank) are used as metrics.
\paragraph{Detection:} To test the detection ability of our model, we choose the candidates with the highest probability as the prediction result and calculate the Hit@1, Hit@2 and Hit@5 as metrics.

To validate our proposed model, we select three baselines for comparison:
\textbf{FUNSD-base} \cite{jaume2019funsd}: The baseline model offered by \citep{jaume2019funsd}. It takes the semantic feature and layout feature of each entity.
\textbf{LayoutLM} \cite{xu2020layoutlm}: It takes the textual contents and layout features as inputs of pretrained model, LayoutLM, to extract text-layout features for each entity.
\textbf{DocStruct} \cite{wang2020docstruct}: It combines multimodal information of textual, layout and visual features of a semantic entity together.

Since we build an end-to-end method to solve the relation extraction task, our inputs are the bounding boxes of single words. However, most existing works only focus on the individual task and take the semantic entities as inputs. The detailed experiment results are shown in Table \ref{tab:entity_link_result}

From the Table \ref{tab:entity_link_result}, we observe that our proposed link prediction model for relation extraction task greatly outperforms the baselines. We attribute the enhancement to the word-wise inputs and downstream model design. Since our inputs are word-wise, our base model can extract more features according to the context of the whole page. The multi-layer transformer in the downstream model can further convert word-level features to the informative feature of entities.

\subsection{Multitask Training Comparison}
The previous experiment results show that our proposed model achieves satisfactory performance on each task. We further compare our proposed model in single-task training and multitask training. The comparison is shown in Table \ref{tab:multitask_result_1} and Table \ref{tab:multitask_result_2}.

\begin{table}[h]
  \centering
  \caption{Comparison Result of Single-task and Multitask Training of Word Grouping Task}
    \begin{tabular}{|l|c|}
    \hline
    \textbf{Word Grouping Task} & \textbf{Accuracy} \bigstrut\\
    \hline
    single-task training & 91.55 \bigstrut\\
    \hline
    multitask training & \textbf{91.94} \bigstrut\\
    \hline
    \end{tabular}%
  \label{tab:multitask_result_1}%
\end{table}%

Table \ref{tab:multitask_result_1} and Table \ref{tab:multitask_result_2} show that multitask training brings improvements to each task. In the relation extraction task, the semantic entity representation is crucial. Because our inputs are word-wise, the word-wise features cannot provide integrated information of an entity. In the word grouping task, however, the labeling process will make the intervals between entities much clearer to refine the entity representation. The teacher forcing rate setting in multitask training also allows the linking prediction model to take the sequence labeling model's result as inputs. In this way, the linking prediction can give feedback, and two tasks reinforce each other.

\section{Conclusion}
In this paper, we proposed an end-to-end multitask method to solve two challenging tasks, word grouping and relation extraction in form understanding. The whole pipeline leveraged advanced neural models, e.g., BERT, Transformer, to extract word features and entity features and utilized BiLSTM and CRF in sequence labeling. For the first time, multitask training is used to solve the two crucial tasks in form understanding. The experiment proved our proposed method's effectiveness in aggregating the words into entities and predicting the relation links between entities in forms. In the future, we will apply our method to other challenging benchmarks and strive to combine more features.

\section{Ethical Considerations}
All the data and techniques used in this paper are open and available online, so we think there are no privacy issues. Moreover, our method is intended to improve form understanding techniques, i.e., converting the unstructured data in forms into structured data for computers. We adopt pretrained language models in our method, and the potential bias can be eliminated thanks to the massive training corpus size.

To sum up, we are motivated to explore state-of-the-art models and methods and create a more automatic life for people.

\bibliographystyle{acl_natbib}
\bibliography{ref}

\clearpage

\section*{Appendix}

\appendix 

\section{Training details}
We train our model on 4 tesla V100 GPUs, and batch size is set as 4 pages. In the text-layout base model, the sliding window size is set as 512, and the stride length is set as 256. The layout feature dimension is set as 128. In the sequence labeling model, the BiLSTM has two layers, and the hidden state size is set as the length of the textual-layout feature. In the link prediction model, the multi-layer transformer has 3 layers, and the negative sampling size is 50.

\section{Metric Explanation}
\subsection{Mean Average Precision (mAP)}
Mean Average Precision is a metric widely used in the area of object detection. It measures the average precision value for different recall value, so the larger mAP is, the better the model performs. 

In our reconstruction task for relation extraction, we detect the key entity for a given entity. For example, the given entity $x$ has $n$ candidates $y_1,y_2,...,y_n$ (the rest entities in the same page) and $m$ of them are the right answers. The recall value can be $\frac{i}{m}$, where $i = 0,1,...,m$. We calculate the biggest precision value for each recall value. We denote the biggest precision when recall equals $\frac{i}{m}$ as $p_i$ and calculate mAP as followed:
\begin{equation}
    mAP = \sum_{i=0}^m p_i * \frac{1}{m}
\end{equation}

\subsection{Mean Rank (mRank)}
Mean Rank is also a metric used in the area of object detection. It measures the average number of wrong answers that rank higher than right answers, so the smaller mRank is, the better the model performs. 

It is easier to think the mRank as the average number of right-wrong reverse pairs in ranking list. For example, our given entity $x$ has $n$ candidates $y_1,y_2,...,y_n$ (the rest entities in the same page). Among the candidates, $m$ of them are right answers and the corresponding indices are $i_1,i_2,...,i_m$ in an ascending order. For the first right answer $y_{i_1}$, the number of wrong answers that rank higher than it is $i_1 - 1$. For the second right answer $y_{i_2}$, the number of wrong answers that rank higher than it is $i_2 - 2$ (excluding $y_{i_1}$ before it). Therefore, we calculate mRank as followed:
\begin{equation}
    mRank = \sum_{k=1}^m i_k - k = \sum_{k=1}^m i_k - \frac{(1+m)m}{2}
\end{equation}

\subsection{Hit@k}
Hit@k is a metric that measures the ratio of cases whose right answers appear in the top $k$ prediction candidates. In our detection task, we detect the key entity for a given entity. We predict the probability for every entity in a page and count the number of entities whose true key entity appears in the top $k$ candidates. We calculate Hit@k through division of the number and the total number of entities. 

\end{document}